# DeepSquare: Boosting the Learning Power of Deep Convolutional Neural Networks with Elementwise Square Operators

Sheng Chen, Xu Wang, Chao Chen, Yifan Lu, Xijin Zhang, Linfu Wen
ByteDance AI Lab
{chensheng.lab, wangxu.ailab, chenchao.cc, luyifan, zhangxijin, wenlinfu}@bytedance.com

## Abstract

*Modern neural network modules which can significantly enhance the learning power usually add too much computational complexity to the original neural networks. In this paper, we pursue very efficient neural network modules which can significantly boost the learning power of deep convolutional neural networks with negligible extra computational cost. We first present both theoretically and experimentally that elementwise square operator has a potential to enhance the learning power of neural networks. Then, we design four types of lightweight modules with elementwise square operators, named as Square-Pooling, Square-Softmin, Square-Excitation, and Square-Encoding. We add our four lightweight modules to Resnet18, Resnet50, and ShuffleNetV2 for better performance in the experiment on ImageNet 2012 dataset. The experimental results show that our modules can bring significant accuracy improvements to the base convolutional neural network models. The performance of our lightweight modules is even comparable to many complicated modules such as bilinear pooling, Squeeze-and-Excitation, and Gather-Excite. Our highly efficient modules are particularly suitable for mobile models. For example, when equipped with a single Square-Pooling module, the top-1 classification accuracy of ShuffleNetV2-0.5x on ImageNet 2012 is absolutely improved by 1.45% with no additional parameters and negligible inference time overhead.*

## 1. Introduction

Deep learning has achieved great success in many areas such as vision, speech, and natural language processing. However, many studies demonstrate that the learning power of deep neural networks (DNNs) heavily depends on their model sizes and computational complexity. On one hand, many researches [1, 2, 3] show that the depth is very important for DNNs. On the other hand, [4] indicates that neural networks should be wide enough to learn disconnected decision regions. Particularly, state-of-the-art convolutional neural network (CNN) architectures such as ResNet [5], DenseNet [6], and AmoebaNet [7] are usually both very deep and wide.

To boost the learning power of deep CNNs, several handcrafted neural network modules have been proposed, for example, bilinear pooling [8], Squeeze-and-Excitation [9], and Gather-Excite [10]. These modern neural network modules usually add too much computational complexity to the original neural networks although they can enhance the learning power a lot. To pursue high efficiency, several carefully tailored CNN architectures have been designed, for example, Xception [11], MobileNetV2 [12], and ShuffleNetV2 [13]. Recently, researchers pay more and more attention to automatic neural network architecture search [7, 14, 15]. The current search methods have a restrict that the search space is the combinations of basic operators and modules. Expectedly, the state-of-the-art architectures found by these search methods cannot surpass the state-of-the-art handcrafted architectures a lot. It is very necessary to enlarge the space of neural network architecture design and search by taking unusual or new basic operators and modules into account.

In this paper, we design four types of lightweight square nonlinear modules with elementwise square (EleSquare for short; see Equation (14) for its mathematical expression) operators to boost the learning power of DNNs (see Figure 1). Our work is initially motivated by [4]. The authors of [4] point out that ReLU activated neural networks cannot produce disconnected decision regions if no layer has more hidden units than the input dimension. When we take EleSquare nonlinear function into account, the neural networks with the same width as above can produce disconnected decision regions (see Figure 2). But there is a trouble that the EleSquare nonlinear function does not satisfy Lipschitz condition. It is very hard to train successfully the DNNs with all activation functions replaced by EleSquare activation functions. Among the quadratic neural network operators or modules, EleSquare operator is the simplest one. A widely studied class of quadratic module is bilinear pooling [8], which is mainly used for fine-grained visual recognition. Bilinear pooling and its variants [16, 17] are not widely used for common visual tasks since they add too much computational complexity to the original neural networks. To solve this problem, we pursue very efficient neural network modules which can significantly boost the learning power of DNNs with negligible extra computational cost. In theory, EleSquare operator is a potential component of the efficient neural network modules. In practice, misusing EleSquare operators probably makes neural networks perform much worse. To get promising new modules with EleSquare operators, we start our design by quickly trying various methods to add EleSquare operators to a simple vanilla CNN model on CIFAR10. Then, four types of efficient modules, named as Square-Pooling, Square-Softmin, Square-Excitation, and Square-Encoding, are carefully designed with EleSquare operators for modern CNN architectures (see Figure 1). In the experiment on ImageNet 2012 dataset, we add our four lightweight modules to Resnet18, Resnet50, and ShuffleNetv2 [13] for better performance. The experimental results show that the learning power of the CNN models can be significantly boosted by our modules. For example, when equipped with a single Square-Pooling module, the top-1 classification accuracy of ShuffleNetV2-0.5x on ImageNet 2012 is absolutely improved by 1.45% with no additional parameters and negligible inference time overhead.

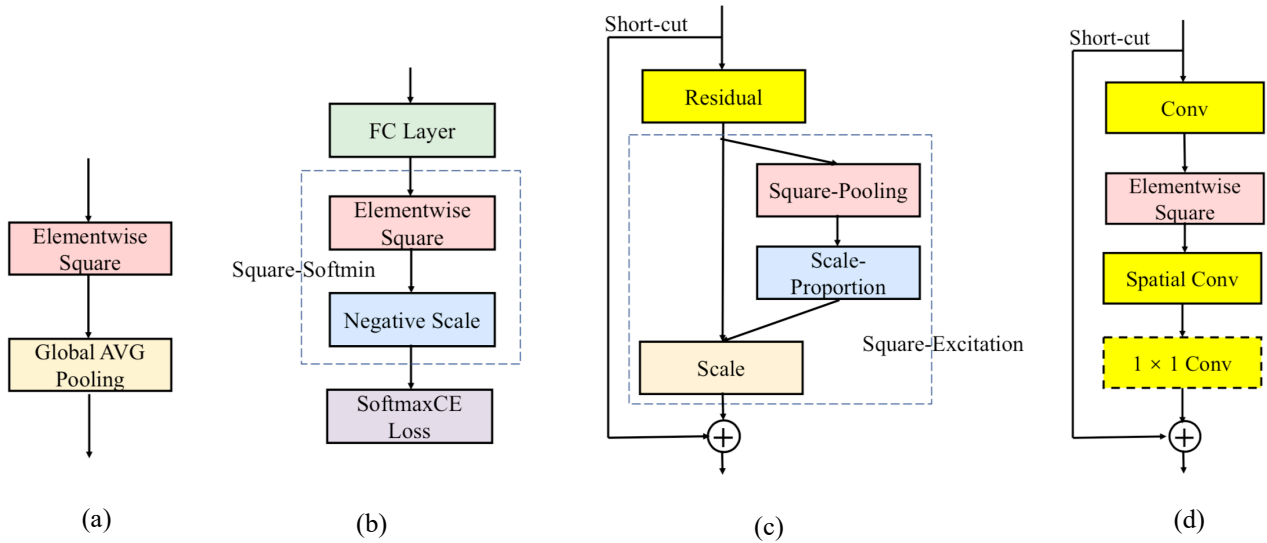

Figure 1: Our four types of DeepSquare modules for modern CNNs. (a) A Square-Pooling module. (b) A Square-Softmin module between the last FC layer and the SoftmaxCE loss. (c) A Square-Excitation module in a residual block. The Scale-Proportion layer is defined by Equation (25). (d) A Square-Encoding module refers to the elementwise square layer before the last spatial convolution of the main branch in a building block. The block is kept invariant to the original except for the additional elementwise square layer.

The main contributions of this paper include:
- We present both theoretically and experimentally that EleSquare operator has a potential to enhance the learning power of DNNs.
- We design four types of lightweight modules with EleSquare operators, named as Square-Pooling, Square-Softmin, Square-Excitation, and Square-Encoding, which bring accuracy improvements even comparable to many complicated modules such as bilinear pooling, Squeeze-and-Excitation, and Gather-Excite.
- Moreover, our highly efficient modules are particularly suitable for mobile models since they only add negligible extra computational cost and number of parameters.

## 2. Related work

Our work is mainly based on EleSquare operator. This is not the first work in which EleSquare operator has been used. EleSquare operator has been used as a small component in many complicated quadratic modules such as Fisher Vector (FV) encoding [18], bilinear pooling [8], and their variants.

**FV encoding**. The FV representation is encoded from a set of local descriptors $x: \{x_t \mid t = 1, \cdots, n\}$ as follows. It computes both the 1-st order statistics

$$\alpha_k = \Omega_k^{-\frac{1}{2}}(x - \mu_k), \quad (1)$$

and the 2-nd order statistics

$$\beta_k = \Omega_k^{-1}(x - \mu_k)^2 - 1, \quad (2)$$

which are aggregated and weighted by the Gaussian mixture model (GMM) posteriors $\gamma$ (see Equation (15) in [18]), with $\mu_k$ and $\Omega_k$ corresponding to the mean and covariance of the $k$-th GMM component respectively (c.f. [8, 18]). Note that the square of a vector should be understood as an elementwise operation. As shown in [18], we can also compute the FV representation by combining the following terms:

$$b_k = \sum_t \gamma_t(k), \quad (3)$$

$$M_k = \sum_t \gamma_t(k) x_t, \quad (4)$$

$$S_k = \sum_t \gamma_t(k) x_t^2. \quad (5)$$

The FVs need to be normalized by the techniques of FV normalizations for better representation. The FV representation is usually a very high-dimensional vector, which should be compressed for compactness.

**Bilinear pooling**. The bilinear pooling forms a global representation by aggregating the location-wise outer-product of two features by global averaging [8, 19]. When the two features are identical, the bilinear pooling is also the second-order pooling [20]. In this case, the second-order statistics

$$B(x) = \frac{1}{n} \sum_{t=1}^{n} x_t x_t^T, \quad (6)$$

is computed as the global representation, which is followed by the elementwise signed square-root ($x \leftarrow sign(x)\sqrt{|x|}$) and the $L2$ normalization ($x \leftarrow x/\|x\|_2$) before it is plugged into linear classifier [8, 19].

Many improved variants [21, 22, 23, 24] of bilinear pooling, second-order pooling have also been proposed in the recent years. When used in CNN models for visual tasks, none of bilinear pooling, second-order pooling and their variants is lightweight.

Our Square-Pooling module is closely related to Lp-pooling [25] with $p = 2$ and generalized-mean (GeM) pooling [26] with the pooling exponent of 2. The Lp-pooling maps the input feature map $I$ to the output $O$ by

$$O = (\sum \sum I(i,j)^p \cdot G(i,j))^{\frac{1}{p}}, \quad (7)$$

where $G$ is a Gaussian kernel (see **2.1** of [25]). The GeM pooling with the pooling exponent of $p$ maps the input feature map $I$ to the output $O$ by

$$O = (\frac{1}{HW} \sum_{i=1}^{H} \sum_{j=1}^{W} I(i,j)^p)^{\frac{1}{p}}, \quad (8)$$

as described in [26].

Our Square-Excitation module is also related to Squeeze-and-Excitation (SE) [9] module and Gather-Excite (GE) [10] module. Both the SE module and the GE module extract contextual information by aggregating feature responses from a large spatial extent and rescale feature channels by a factor mapped from the contextual information. The SE module has two fully connected (FC) layers which contain many learnable parameters. The GE module is cheaper than the SE module, both in terms of number of added parameters and computational complexity. However, the GE module is still not lightweight enough especially for mobile modules since it contains many exponential operations in its sigmoid operators.

## 3. The potential of EleSquare operator

### 3.1. The situation when neural networks are not wide enough

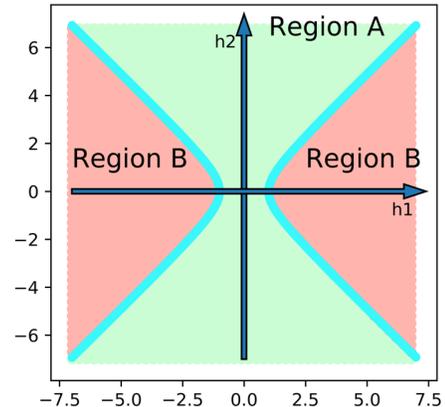

Figure 2: Hyperbolic decision boundary and regions. Region A and B correspond to the two decision regions of the binary classification. Region B consists of two disconnected regions.

Both depth and width of DNNs are very important for the learning power of the models. Here we consider the learning power of neural networks when they are not wide enough. [4] points out that ReLU activated neural networks cannot produce disconnected decision regions if no layer has more hidden units than the input dimension. We note that this is not true for the neural networks activated by the EleSquare functions. Without loss of generality, we analyze the case of the neural network for binary classification with only two fully connected (FC) layers and with two-dimensional input and two hidden units. Let $\sigma : \mathbb{R} \to \mathbb{R}$ be the activation function of the neural network which elementwise applied on the hidden units

$$h = \begin{pmatrix} h_1 \\ h_2 \end{pmatrix} = f_1(x), \quad (9)$$

where $x \in \mathbb{R}^2$ is the input vector and $f_1$ is the affine transformation corresponding to the first FC layer. The output $y$ of the second FC layer can be computed as

$$y = \begin{pmatrix} y_1 \\ y_2 \end{pmatrix} = W^T \sigma(h) + b, \quad (10)$$

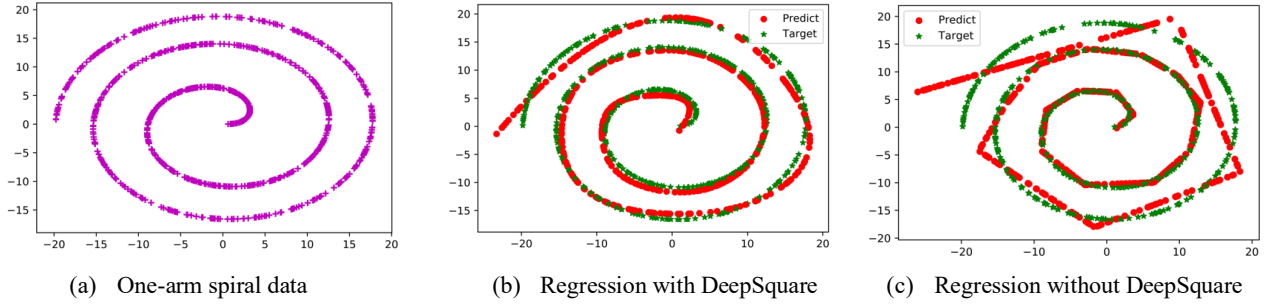

(a) One-arm spiral data  (b) Regression with DeepSquare  (c) Regression without DeepSquare

Figure 3: Regression w/o DeepSquare on one-arm spiral data.

where

$$W = \begin{pmatrix} w_{11} & w_{12} \\ w_{21} & w_{22} \end{pmatrix} \in \mathbb{R}^{2 \times 2} \quad (11)$$

and

$$b = \begin{pmatrix} b_1 \\ b_2 \end{pmatrix} \in \mathbb{R}^2 \quad (12)$$

are the weight matrix and the bias of the second FC layer respectively. When the training loss of the neural network is cross-entropy with softmax (SoftmaxCE), the decision boundary is determined by solving equation

$$y_1 = y_2. \quad (13)$$

When the activation function $\sigma$ is ReLU, paper [4] has proven that the neural network cannot produce disconnected decision regions. When $\sigma$ is the EleSquare function, i.e.,

$$\sigma(t) = t^2 \quad \text{for } t \in \mathbb{R}, \quad (14)$$

Equation (13) is equivalent to the equation

$$w_{11}h_1^2 + w_{21}h_2^2 + b_1 = w_{12}h_1^2 + w_{22}h_2^2 + b_2, \quad (15)$$

which can be simplified as

$$(w_{11} - w_{12})h_1^2 + (w_{21} - w_{22})h_2^2 = b_2 - b_1. \quad (16)$$

When

$$(w_{11} - w_{12})(w_{21} - w_{22}) < 0 \quad (17)$$

and

$$b_2 - b_1 \neq 0, \quad (18)$$

the decision boundary determined by Equation (16) is a hyperbola which can produce disconnected decision regions (see Figure 2). On one hand, the above analysis indicates that EleSquare operator has a potential to enhance the learning power of DNNs in some cases. On the other hand, EleSquare operator does not satisfy Lipschitz condition, so it is not recommended as the activation function of very deep neural networks. Nevertheless, we can fetch more benefits with kinds of combinations of EleSquare operators and other operators. In the following of this paper, the various ways to use EleSquare operator in DNNs are indiscriminately called DeepSquare unless otherwise specified.

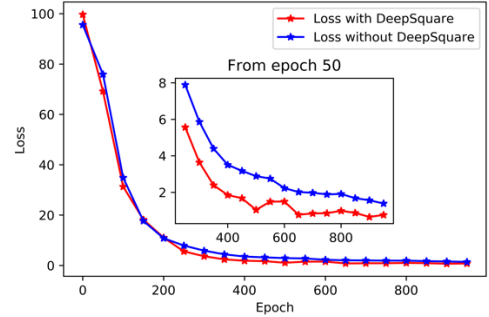

Figure 4: Test regression loss of mean squared error

### 3.2. Smoother fitting with DeepSquare

There is a natural benefit that the function smoothness of the neural networks activated by ReLU can be enhanced by DeepSquare. ReLU function is not smooth since its derived function is not continuous. Let ReLU-Square denote the composition of ReLU and EleSquare functions, i.e.,

$$\sigma(t) = (max(0, t))^2 \quad \text{for } t \in \mathbb{R}. \quad (19)$$

It is clear that the derived function of ReLU-Square is continuous. Therefore, ReLU-Square is smoother than ReLU and the former results in smoother neural networks.

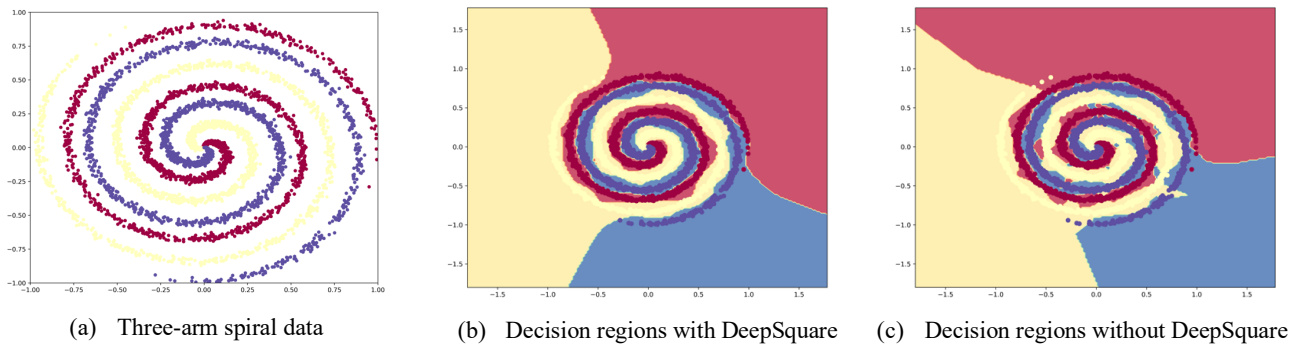

(a) Three-arm spiral data     (b) Decision regions with DeepSquare     (c) Decision regions without DeepSquare

Figure 5: Classification w/o DeepSquare on three-arm spiral data.

Smoother fitting with DeepSquare can lead to better performance in some regression and classification tasks, as shown in our numerical experiments.

**Regression on one-arm spiral data.** Here we consider the one-arm spiral data (see Figure 3(a)) as the trajectory corresponding to the map from a time point to a data-space point. We use a two-FC-layer neural network with the hidden dimension of 32 and the output dimension of 2 to regress this map. For the case with DeepSquare, the hidden units are activated by ReLU-Square. For the case without DeepSquare, the hidden units are activated by ReLU. The regressed trajectories with and without DeepSquare are shown in Figure 3(b) and Figure 3(c) respectively. It is shown that the regressed trajectory with DeepSquare is much smoother than that without DeepSquare. The regression losses with and without DeepSquare on the test data are shown in Figure 4. These results demonstrate that DeepSquare significantly improves the regression performance on the one-arm spiral data.

**Classification on three-arm spiral data.** For the three-arm spiral data (see Figure 5(a)), we use a two-FC-layer neural network with the hidden dimension of 32 and the output dimension of 3 to tell the three arms apart. The setting of the activation functions is the same to the above regression experiment. The decision regions of the trained classification models with and without DeepSquare are shown in Figure 5(b) and Figure 5(c) respectively. It is shown that the decision boundaries with DeepSquare are smoother than those without DeepSquare. The classification losses and accuracies with/without DeepSquare on test data are shown in Figure 6. The classification performance on the three-arm spiral data is also impoved by DeepSquare.

Note that some nonlinear functions of an exponential form, such as softplus, tanh and sigmoid, can also make the neural networks smoother, but they have much more computational complexity than the composition of ReLU and EleSquare functions.

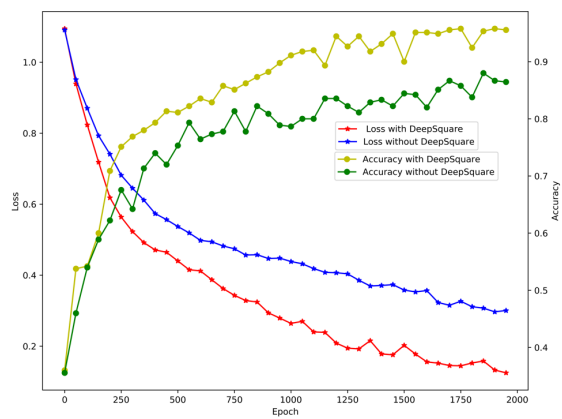

Figure 6: Test loss and accuracy of classification

## 4. DeepSquare for CNNs

### 4.1. A glance at DeepSquare for a vanilla CNN on CIFAR10

Despite its big potential, DeepSquare needs to be used carefully in practical models. Casually using DeepSquare may cause performance degradation or model divergence (see Table 2). By taking a quick glance at various usages of DeepSquare in a vanilla CNN classification model on CIFAR10, we will get some guidelines of how to use DeepSquare properly in the following. The vanilla CNN backbone to which we will apply DeepSquare is shown in Table 1. Among different methods of DeepSquare, which add one or several EleSquare operators to different positions of the CNN, we make a performance comparison in Table 2. Although there are much more methods to apply DeepSquare than those we listed, few of them can achieve improved classification accuracy on CIFAR10. DeepSquare3, which only adds an EleSquare operator before the GAP layer, achieves 1.06% absolute improvement of accuracy compared to the original model. DeepSquare5s-, which adds an EleSquare operator and a negative scale layer after the last FC layer, achieves 1.05% absolute improvement of accuracy compared to the original model. In the rest of this paper, the combination of EleSquare and GAP as DeepSquare3 is called Square-Pooling (see Figure 1(a)), and the combination of EleSquare and negative scale layer as DeepSquare5s- is called Square-Softmin when used between the last FC layer and SoftmaxCE (see Figure 1(b)).

| Input size | Layer ID | Layer type | Output size |
|---|---|---|---|
| 32×32×3 | 1 | Conv3×3s2 | 16×16×32 |
| 16×16×32 | 2 | Conv3×3s2 | 8×8×64 |
| 8×8×64 | 3 | Conv3×3s2 | 4×4×128 |
| 4×4×128 | 4 | GAP | 1×1×128 |
| 128 | 5 | FC | 10 |

Table 1. The vanilla CNN backbone without DeepSquare. Conv3×3s2 denotes the convolution layer with kernel size 3×3 and stride 2, where the Batch Normalization layer and ReLU layer is also applied.

| Method | Position of EleSquare | CIFAR10 acc. (%) | Improvement (%) |
|---|---|---|---|
| Original | None | 81.71 | 0 |
| DeepSquare1 | 1_2 | 80.32 | -1.39 |
| DeepSquare2 | 2_3 | 81.69 | -0.02 |
| DeepSquare3 | 3_4 | **82.77** | **1.06** |
| DeepSquare4 | 4_5 | 81.39 | -0.32 |
| DeepSquare5+ | 5_ | 76.88 | -4.83 |
| DeepSquare5- | 5_ | 82.24 | 0.53 |
| DeepSquare5s+ | 5_ | 81.93 | 0.22 |
| DeepSquare5s- | 5_ | **82.76** | **1.05** |
| DeepSquare6 | 1_2_3_4, 5_ | divergence | - |
| DeepSquare7 | 1_2_3_4 | 79.75 | -1.96 |
| DeepSquare8 | 3_4, 5_ | 82.68 | 0.97 |

Table 2. Performance comparison among different methods of DeepSquare applied to the vanilla CNN. Position 1_2 of EleSquare refers to the method of adding EleSquare between Layer 1 and Layer 2 of the CNN. DeepSquare5- refers to applying minus EleSquare after Layer 5.

## 4.2. A closer look at Square-Pooling and Square-Softmin

**Square-Pooling.** A Square-Pooling module maps the input $F$ of size $H \times W \times C$ to the output $G$ of size $C$. The map is computed as:

$$G_k = \frac{1}{H \cdot W} \sum_{i,j} F_{i,j,k}^2, \quad (20)$$

where $G_k$ is the $k$-th element of $G$ and $F_{i,j,k}$ is the element of spatial position $(i,j)$ in the $k$-th channel of $F$. In statistics, Square-Pooling computes the second order origin moment over all spatial positions for each channel of the input. Moreover, (20) can be rewritten as:

$$G_k = M_k^2 + S_k^2, \quad (21)$$

where $M_k$ and $S_k^2$ are the mean and variance of all elements in the $k$-th channel of $F$ respectively:

$$M_k = \frac{1}{H \cdot W} \sum_{i,j} F_{i,j,k}, \quad (22)$$

$$S_k^2 = \frac{1}{H \cdot W} \sum_{i,j} (F_{i,j,k} - M_k)^2, \quad (23)$$

By Equation (21), we can infer that Square-Pooling extracts a more discriminative feature vector than GAP in usual cases since GAP only takes $(M_k)_{1 \leq k \leq C}$ as the feature vector.

Square-Pooling is very near to GeM [26] with the pooling exponent of 2. A GeM (p=2) module can be regarded as the combination of a Square-Pooling module and an elementwise square-root operator. Compared with GeM (p=2), Square-Pooling can achieve competitive or better performance with less computational complexity, as shown in Table 3. We also compare Square-Pooling with other variants which compute higher order origin moments in Table 3. These variants achieve inferior performance over Square-Pooling as expected.

| Method | Order of moments | CIFAR10 acc. (%) | Improvement (%) |
|---|---|---|---|
| Original | 1 | 81.71 | 0 |
| GeM (p=2) | - | 82.51 | 0.8 |
| **Square-Pooling** | 2 | **82.77** | **1.06** |
| Variant1 | 3 | 82.23 | 0.52 |
| Variant2 | 4 | 78.84 | -2.87 |
| Variant3 | 5 | 80.34 | -1.37 |
| Variant4 | 6 | 73.00 | -8.71 |

Table 3. Performance comparison among Square-Pooling, GeM (p=2) and other variants. Variant1 computes the 3-rd order origin moment over all spatial positions of each channel of the input.

**Square-Softmin.** We use a Square-Softmin module after the last FC layer to produce nonpositive logits. Let $x$ be the output vector of the last FC layer. When it is not followed by Square-Softmin, its elements are the logits for SoftmaxCE. When Square-Softmin is used, it maps $x$ to the final logits as:

$$y_k = -s_k x_k^2 = -(\sqrt{s_k} x_k)^2, \quad (24)$$

where $x_k$ is the $k$-th element of $x$ and $s_k$ is a nonnegative learnable per-channel scaling factor. In some cases, all the scaling factors of the Square-Softmin module are set to share the same learnable parameter to reduce the overfitting risk. Note that the absolute values of all the scaling factors can be merged into the weight parameters of the last FC layer in the test phase. Since the logits produced by Square-Softmin are nonpositive, the maximal logit is the closest to zero with the minimal absolute value. In this sense, we give the nickname Softmin to the softmax operator following a Square-Softmin module.

If we change the positive or negative signs of some elements of $x$ in Equation (24), the logits produced by Square-Softmin will be invariant. Therefore, the decision regions of the network with Square-Softmin are more symmetrical than those without Square-Softmin.

### 4.3. More DeepSquare modules for modern CNNs

DeepSquare modules more than Square-Pooling and Square-Softmin can be used to boost the performance of modern CNNs, such as ResNet and ShuffleNetV2, since modern CNNs usually have many building blocks.

**Square-Excitation**. Inspired by SE module and GE module, we design a Square-Excitation module for each building block of a modern CNN. Let $F$ be the output feature map of a main branch of a block. Passing $F$ to a Square-Pooling module results in a vector $G$ by Equation (20). The Square-Excitation module rescales each channel of $F$ according to $G$ before $F$ is added or concatenated to the short-cut branch (see Figure 1(c)). The rescaling factor for the $k$-th channel of $F$ is computed as:

$$s_k = \frac{G_k}{G_k + \alpha^2}, \quad (25)$$

where $G_k$ is the $k$-th element of the $G$ and $\alpha$ is a learnable parameter shared by all channels. We give a short name Scale-Proportion to the layer defined by Equation (25). Note that, unlike SE module and GE module, Square-Excitation does not use any exponential nonlinearity such as sigmoid and tanh.

**Square-Encoding**. Another DeepSquare module is Square-Encoding which uses an EleSquare operator before the last spatial convolution layer in the main branch of each building block (see Figure 1(d)). Square-Encoding makes the spatial convolution encode new features partly similar to FV encoding (see Equation (5)). Without soft assignments or posteriors like FV encoding, Square-Encoding gets the weights of the square terms by learning the weights of the spatial convolution.

The effectiveness of Square-Pooling, Square-Softmin, Square-Excitation, and Square-Encoding for modern CNNs will be demonstrated in the image classification experiments on ImageNet 2012 [27] in the next section.

## 5. Experiment

In this section, we use ResNet18, ResNet50, and ShuffleNetV2 as the base CNNs to perform our experiments on ImageNet 2012 classification dataset. Firstly, based on ResNet18, we check the effectiveness of Square-Pooling, Square-Softmin, Square-Excitation, Square-Encoding, and several combinations of the four DeepSquare modules. Then, based on ResNet50, we evaluate the performance of several typical DeepSquare methods for big CNN models. Finally, based on ShuffleNetV2, we demonstrate the effectiveness of our methods for mobile CNN models.

**Experimental setup**. Pytorch [28] framework is used to train our CNN models on ImageNet 2012 with 8 Nvidia Tesla V100 GPUs. The standard ResNet data augmentation [5] is used to preprocess the training images. DALI [29] library is used to speed up the data preprocessing. The optimization method is stochastic gradient descent (SGD) with Nesterov momentum, and the momentum is 0.9. The total mini-batch size is 1024, and the initial learning rate is 0.4. In the first 5 epochs, we linearly increase the learning rate from 0 to the initial learning rate for warmup. After the warmup, we use the cosine learning rate decaying schedule suggested by [30]. For the CNNs based on ResNet18 and ResNet50, the weight decay is $1 \times 10^{-4}$ and the number of training epochs is 120. For the CNNs based on ShuffleNetV2, the weight decay is $4 \times 10^{-5}$ and the number of training epochs is 400. To reduce the overfitting risk, the dropout ratio of 0.2 is used for training the CNNs with DeepSquare and all the scaling factors of the Square-Softmin module are set to share the same learnable parameter. In the test phase, the classification accuracy is evaluated on ImageNet 2012 validation set with the single 224×224 center crop from the resized image whose shorter side is 256.

| Method | Top-1 acc. (%) | Top-5 acc. (%) |
|---|---|---|
| ResNet18 (our impl.) | 70.98 | 89.92 |
| Square-Pooling (SP) | 71.41 | 90.02 |
| Square-Softmin (SS) | 71.21 | 90.22 |
| Square-Softmin (200 epochs) | **71.92** | **90.39** |
| Square-Excitation (SEx) | 71.38 | 90.12 |
| Square-Encoding (SEn) | 71.56 | 90.25 |
| SP & SEx | 71.60 | 90.17 |
| SP & SEn | 70.40 | 89.52 |
| SP & SEx & SEn | 70.74 | 89.60 |

Table 4. Comparison of different DeepSquare methods with a ResNet18 baseline over classification accuracy on the ImageNet 2012 validation set (single 224×224 center crop).

**DeepSquare methods on ResNet18.** Square-Pooling, Square-Softmin, Square-Excitation and Square-Encoding can effectively boost the classification accuracy of ResNet18 as shown in Table 4. These modules can provide even more benefit when a longer training schedule is used. For example, ResNet18 with Square-Softmin can achieve 71.92% top-1 accuracy on ImageNet 2012 if the number of training epochs increases to 200. There are too many possible combinations of the four DeepSquare modules. We only show several typical combinations in Table 4. The combination (SP & SEx) of Square-Pooling and Square-Excitation achieves better accuracy than Square-Pooling or Square-Excitation. However, many other combinations get worse results.

| Method | Top-1 acc. (%) | Top-5 acc. (%) |
|---|---|---|
| ResNet50 (our impl.) | 77.06 | 93.34 |
| Square-Pooling (SP) | 77.39 | 93.55 |
| Square-Softmin (SS) | 77.25 | 93.32 |
| Square-Encoding (SEn) | 77.35 | 93.55 |
| **SP & SEx** | **78.14** | **94.05** |

Table 5. Comparison of different DeepSquare methods with a ResNet50 baseline over classification accuracy on the ImageNet 2012 validation set (single 224×224 center crop).

**DeepSquare methods on ResNet50.** We show the performance of several typical DeepSquare methods based on ResNet50 in Table 5. These methods achieve improved accuracy again. Especially, SP & SEx, i.e., the combination of Square-Pooling and Square-Excitation achieves the best top-1 accuracy of 78.14% among these methods. We also make a performance comparison between SP & SEx and other related methods based on ResNet50 in Table 6. Compared to the state-of-the-art methods, SP & SEx achieves competitive performance with negligible additional parameters and the minimum inference time overhead.

| Method | +Parameters | Top-1 acc. (%) | Top-5 acc. (%) |
|---|---|---|---|
| **SP & SEx** (ours) | **0%** | 78.14 | 94.05 |
| MPN-COV [23] | 120% | 77.26 | 93.46 |
| iSQRT-COV [31] | 120% | 77.86 | 93.78 |
| GSoP-Net1 [32] | 10% | 77.68 | 93.98 |
| GSoP-Net2 [32] | 130% | **78.81** | **94.36** |
| SE [9] | 10% | 76.71 | 93.38 |
| GE-$\theta$ [10] | 22% | 78.00 | 94.13 |

Table 6. Performance comparison between SP & SEx and other related methods based on ResNet50 over classification accuracy on the ImageNet 2012 validation set (single 224×224 crop).

**DeepSquare methods on ShuffleNetV2.** To verify the effectiveness of our methods for mobile CNN models, we also evaluate the performance of several typical DeepSquare methods based on ShuffleNetV2. As shown in Table 7, our DeepSquare methods can effectively boost the learning power of ShuffleNetV2.

| Method | Top-1 acc. (%) | Top-5 acc. (%) |
|---|---|---|
| ShuffleNetV2 [13] | 69.4 | - |
| ShuffleNetV2 (our impl.) | 68.89 | 88.58 |
| Square-Pooling (SP) | 70.53 | 89.49 |
| Square-Softmin (SS) | 69.78 | 89.04 |
| Square-Encoding (SEn) | 69.58 | 89.01 |
| **SP & SEx** | **70.76** | **89.63** |
| ShuffleNetV2-0.5× [13] | 60.3 | - |
| ShuffleNetV2-0.5× (our impl.) | 59.94 | 81.82 |
| ShuffleNetV2-0.5× with SP | **61.39** | **82.98** |

Table 7. Performance of several typical DeepSquare methods based on ShuffleNetV2 for the ImageNet 2012 (test on the single 224×224 center crop).

## 6. Conclusion

In this paper, we presented both theoretically and experimentally that elementwise square operator has a big potential to boost the learning power of neural networks. We designed four types of lightweight modules with elementwise square operators, named as Square-Pooling, Square-Softmin, Square-Excitation, and Square-Encoding. Our experiment on ImageNet 2012 dataset show that our modules can bring significant accuracy improvements to the base CNN models. Our lightweight modules can bring accuracy improvements even comparable to many complicated modules such as bilinear pooling, Squeeze-and-Excitation, and Gather-Excite. Our highly efficient modules are particularly suitable for mobile models since they only add negligible extra computational cost and number of parameters. We hope that the space of neural network architecture design and search can be further enlarged by taking elementwise square operator into account after our work.

## References

[1] M. Telgarsky. Benefits of depth in neural networks. In COLT, 2016.
[2] R. Eldan, and O. Shamir. The power of depth for feedforward neural networks. In COLT, 2016.
[3] M. Raghu, B. Poole, J. Kleinberg, S. Ganguli, and J. Sohl-Dickstein. On the expressive power of deep neural networks. In ICML, 2017.

[4] Q. Nguyen, M. Mukkamala, and M. Hein. Neural networks should be wide enough to learn disconnected decision regions. In ICML, 2018.

[5] K. He, X. Zhang, S. Ren, and J. Sun. Deep residual learning for image recognition. In CVPR, 2016.

[6] G. Huang, Z. Liu, L. van der Maaten, and K. Q. Weinberger. Densely connected convolutional networks. In CVPR, 2017.

[7] E. Real, A. Aggarwal, Y. Huang, and Q. V. Le. Regularized Evolution for Image Classifier Architecture Search. In AAAI, 2019.

[8] T.-Y. Lin, A. RoyChowdhury, and S. Maji. Bilinear CNN models for fine-grained visual recognition. In ICCV, 2015.

[9] J. Hu, L. Shen, and G. Sun. Squeeze-and-excitation networks. In CVPR, 2018.

[10] J. Hu, L. Shen, S. Albanie, G. Sun, and A. Vedaldi. Gather-excite: Exploiting feature context in convolutional neural networks. In NIPS, 2018.

[11] F. Chollet. Xception: Deep learning with depthwise separable convolutions. In CVPR, 2017.

[12] M. Sandler, A. Howard, M. Zhu, A. Zhmoginov, and L.-C. Chen. MobileNetV2: Inverted residuals and linear bottlenecks. In CVPR, 2018.

[13] N. Ma, X. Zhang, H.-T. Zheng, and J. Sun. Shufflenet V2: Practical guidelines for efficient CNN architecture design. In ECCV, 2018.

[14] M. Tan, B. Chen, R. Pang, V. Vasudevan, and Q. V. Le. Mnasnet: Platform-aware neural architecture search for mobile. arXiv preprint arXiv:1807.11626, 2018.

[15] B. Wu, X. Dai, P. Zhang, Y. Wang, F. Sun, Y. Wu, Y. Tian, P. Vajda, Y. Jia, and K. Keutzer. FBNet: Hardware-aware efficient convnet design via differentiable neural architecture search. arXiv preprint arXiv:1812.03443, 2018.

[16] S. Kong, C. Fowlkes. Low-Rank Bilinear Pooling for Fine-Grained Classification. In CVPR, 2017.

[17] K. Yu, and M. Salzmann. Statistically-motivated Second-order Pooling. In ECCV, 2018.

[18] J. Sánchez, F. Perronnin, T. Mensink, and J. Verbeek. Image classification with the fisher vector: Theory and practice. International journal of computer vision, 105(3):222–245, 2013.

[19] T.Y. Lin, S. Maji. Improved Bilinear Pooling with CNNs. In BMVC, 2017.

[20] J. Carreira, R. Caseiro, J. Batista, and C. Sminchisescu. Semantic segmentation with second-order pooling. In European Conference on Computer Vision (ECCV), 2012.

[21] Y. Gao, O. Beijbom, N. Zhang, and T. Darrell. Compact Bilinear Pooling. In CVPR, 2016.

[22] Y. Cui, F. Zhou, J. Wang, X. Liu, Y. Lin, and S. Belongie. Kernel Pooling for Convolutional Neural Networks. In CVPR, 2017.

[23] P. Li, J. Xie, Q. Wang, and W. Zuo. Is Second-Order Information Helpful for Large- Scale Visual Recognition? In ICCV, 2017.

[24] K. Yu, M. Salzmann. Sutatistically-motivated Second-order Pooling. In ECCV, 2018.

[25] P. Sermanet, S. Chintala, Y. LeCun. Convolutional neural networks applied to house numbers digit classification. In ICPR, 2012.

[26] F. Radenović, G. Tolias, and O. Chum. Fine-tuning cnn image retrieval with no human annotation. TPAMI, 2018.

[27] O. Russakovsky, J. Deng, H. Su, J. Krause, S. Satheesh, S. Ma, Z. Huang, A. Karpathy, A. Khosla, M. Bernstein, A. C. Berg, and L. Fei-Fei. ImageNet Large Scale Visual Recognition Challenge. IJCV, 2015.

[28] A. Paszke, S. Gross, S. Chintala, G. Chanan, E. Yang, Z. DeVito, Z. Lin, A. Desmaison, L. Antiga, and A. Lerer. Automatic differentiation in PyTorch. In NIPS Autodiff Workshop: The Future of Gradient-based Machine Learning Software and Techniques, 2017.

[29] T. Gale, P. Tredak, S. Layton, A. Ivanov, and S. Panev. https://github.com/NVIDIA/DALI, 2018

[30] T. He, Z. Zhang, H. Zhang, Z. Zhang, J. Xie, and M. Li. Bag of tricks for image classification with convolutional neural networks. arXiv preprint arXiv:1812.01187, 2018.

[31] P. Li, J. Xie, Q. Wang, and Z. Gao. Towards faster training of global covariance pooling networks by iterative matrix square root normalization. In CVPR, 2018.

[32] Z. Gao, J. Xie, Q. Wang, and P. Li. Global Second-order Pooling Convolutional Networks. arXiv preprint arXiv:1811.12006, 2018.